\documentclass[vecarrow]{svmult}
\usepackage{mathptmx}
\usepackage{helvet}
\usepackage{courier}
\usepackage{makeidx}
\usepackage{amsmath,amsfonts}
\usepackage{graphicx}
\usepackage{cite}
\usepackage[bottom]{footmisc}
\usepackage{bm}

\newtheorem{assumption}{Assumption}
\begin{document}

\title*{Analytically Informed Inverse Kinematics Solution at Singularities}
\author{Andreas M\"uller}
\institute{  Andreas M\"uller\at
  Johannes Kepler University, Linz, Austria,
  \email{a.mueller@jku.at}}
\titlerunning{AI-IK --- Analytically Informed Inverse Kinematics Solution at Singularities}
\maketitle
\vspace{-12ex}

\abstract{Near kinematic singularities of a serial manipulator, the inverse kinematics
(IK) problem becomes ill-conditioned, which poses computational problems for
the numerical solution. Computational methods to tackle this issue are based
on various forms of a pseudoinverse (PI) solution to the velocity IK
problem. The damped least squares (DLS) method provides a robust solution
with controllable convergence rate. However, at singularities, it may not
even be possible to solve the IK problem using any PI solution when certain
end-effector motions are prescribed. To overcome this problem, an
analytically informed inverse kinematics (AI-IK) method is proposed. The key
step of the method is an explicit description of the tangent aspect of
singular motions (the analytic part) to deduce a perturbation that yields a
regular configuration. The latter serves as start configuration for the
iterative solution (the numeric part). Numerical results are reported for a
7-DOF Kuka iiwa.}

\keywords{Inverse kinematics, singularities, pseudoinverse, damped pseudoinverse, Newton--Raphson method, higher-order analysis, kinematic tangent cone.}

\parindent0pt

\section{Introduction\vspace{-2ex}}

The numerical solution of the inverse kinematics (IK) problem of serial
robotic manipulators near singularities has been a topic for many years. The
damped least squares (DLS) method was introduced \cite{Wampler_DPI_1986} as
a computationally efficient method that is robust against singularities. Its
robustness comes at the expense that the convergence rate, and thus the
tracking accuracy, depends on a damping parameter. Various strategies for
adaptively selecting the damping factor were proposed \cite%
{ChanLawrence1988,NakamuraHanafusa1986,MayorgaWongMilano1992,Chiaverini1997}
that are all based on some measure of distance to a singularity. A problem,
which has not been addressed adequately in the literature, is that the DLS
method, as well as other methods relying on a pseudoinverse (PI), may get
stuck when a robot that is in a singularity is commanded to execute
instantaneous EE motions that are not in the image space of the manipulator
Jacobian. One reference where this was addressed is \cite{DeoWalker1993},
where an IK solution is introduced that takes into account the second order
kinematics, i.e. the Hessian in addition to the Jacobian. In this paper, a
method for solving the IK problem at singularities is introduced that does
not suffer from this lock-up phenomenon. The key element of the method is an
initial regularization of the Jacobian by means of a perturbation that will
lead to nearby a regular configuration. This perturbation is deduced from an
analytic description of possible singular motions through the singularity.
The so obtained regularized configuration is used as initial value of the IK
problem that can be solved with an undamped pseudoinverse method. Numerical
results are reported for a 7-DOF Kuka LBR iiwa.

\section{The Inverse Kinematics Problem%
\vspace{-2ex}}

Denote with ${\mathbb{V}}^{n}$ the joint space of a serial robot. The
forward kinematics map $f:{\mathbb{V}}^{n}\rightarrow W$ assigns to a joint
coordinate vector $\mathbf{q}\in {\mathbb{V}}^{n}$ an end-effector (EE) pose 
$\mathbf{C}\in W$ in workspace. The latter is a subspace $W\subset SE\left(
3\right) $ of dimension $m$, according to the task. It is assumed in the
following that $m\leq n$. If $m<n$, the manipulator is \emph{kinematically
redundant}. The EE-velocity $\mathbf{V}\in \mathfrak{w}$ is determined by
the joint velocity $\dot{\mathbf{q}}\in {\mathbb{R}}^{n}$ as $\mathbf{V}=%
\mathbf{J}\left( \mathbf{q}\right) \dot{\mathbf{q}}$, where $\mathbf{J}$ is
the \emph{geometric Jacobian}. Let $m$ be the maximal rank of the
FK-Jacobian, and $r:=\mathrm{rank}~\mathbf{J}\left( \mathbf{q}\right) $. A
configuration $\mathbf{q}\in {\mathbb{V}}^{n}$ where $r<m$ are \emph{forward
kinematics singularities} of the manipulator. Motions with $r<m$ are called 
\emph{singular motions}.

The \emph{geometric IK problem} is to compute a joint coordinate vector for
given EE-pose, i.e. to solve $\mathbf{q}=f^{-1}\left( \mathbf{C}\right) $.
This problem has no unique solution, for non-redundant as well redundant
robots. Closed form solutions are known for special cases only \cite%
{RaghavanRoth1993,HustyPfurnerSchroecker2007}. The \emph{velocity IK problem}
is to compute the joint velocity $\dot{\mathbf{q}}$ for given EE velocity $%
\mathbf{V}$, at a given configuration $\mathbf{q}$. A solution is given in
terms of the weighted pseudoinverse as $\dot{\mathbf{q}}=$ $\mathbf{J}_{%
\mathbf{M}}^{+}\mathbf{V}$. Here the weight $\mathbf{M}$ serves as a metric
on ${\mathbb{V}}^{n}$ to homogenize different units in $\mathbf{q}$. This
solution gives rise to a solution of the geometric IK problem. Best know are
the CLIK algorithms \cite{SicilianoCLIC1990}, which combine numerical
integration of the velocity relation with a regulator feedback to avoid
drift. It thus relies on the (pseudo)inverse solution. The velocity IK
solution also gives rise to iterative solution with a Newton--Raphson method.

When approaching a singularity the Jacobian becomes ill-conditioned.
Kinematically this implies that small EE motions lead to large joint
motions. Mathematically, a solution can be obtained that produces these
(undesirable) motions, e.g. using SVD. To alleviate the problem of large
motions, in \cite{NakamuraHanafusa1986,Wampler_DPI_1986} the IK problem is
amended by adding an additional damping term (thus the robot kinematics is
changed), which leads to the damped least squares (DLS) method (see Sect. %
\ref{secDLS}). 
\begin{figure}[tbp]
\hfill a)\includegraphics[height=7cm]{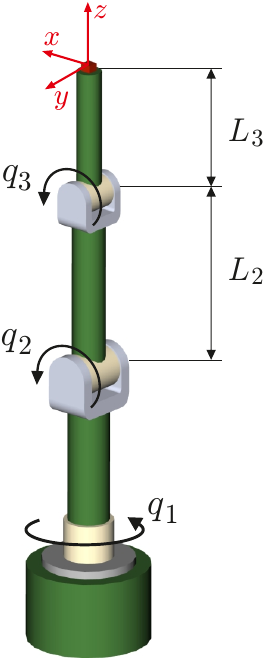} \hfill b)%
\includegraphics[height=7cm]{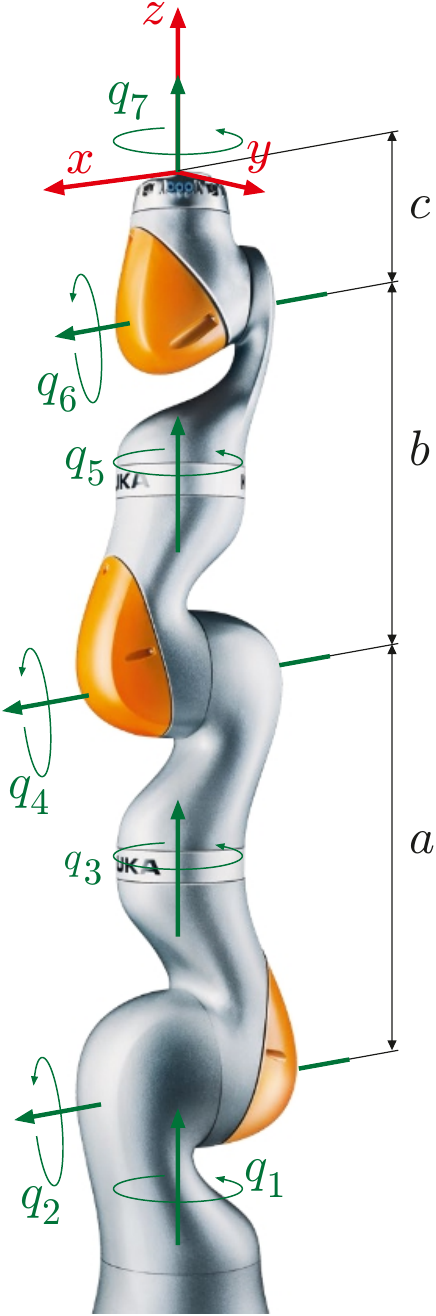} \hfill\ 
\caption{a) 3R regional robot and b) 7-DOF Kuka iiwa 14 R820, both in a
singular configuration.}
\label{figExamples}
\end{figure}

While the DLS provides a robust solution (despite the need for tuning the
damping parameter $\lambda $), it may not allow for solving the IK problem
when the desired EE motion can instantaneous not be produced by joint
motions. A simple example is the 3R regional robot in Fig. \ref{figExamples}%
a), which is in a singularity $\mathbf{q}_{0}$. At this singular
configuration, the Jacobian and its pseudoinverse are%
\begin{equation}
\mathbf{J}\left( \mathbf{q}_{0}\right) =\left( 
\begin{array}{ccc}
0 & {L_{3}}+{L_{2}} & {L_{3}} \\ 
0 & 0 & 0 \\ 
0 & 0 & 0%
\end{array}%
\right) ,\ \ \mathbf{J}^{+}\left( \mathbf{q}_{0}\right) =\left( 
\begin{array}{ccc}
0 & 0 & 0 \\ 
\frac{{L_{3}}+{L_{2}}}{{L_{3}^{2}}+({L_{3}}+{L_{2}})^{2}} & 0 & 0 \\ 
\frac{{L_{1}}}{{L}_{3}^{2}+(L_{3}+L_{2})^{2}} & 0 & 0%
\end{array}%
\right) .  \label{J3R}
\end{equation}%
Clearly, if the EE is commanded to move down in the $y$-$z$-plane, which
corresponds to a prescribed velocity $\mathbf{V}=\left( 0,v_{y},v_{z}\right) 
$, the joint velocity computed from the IK is $\dot{\mathbf{q}}=\mathbf{0}$.
Thus no solution to the geometric IK problem would be obtained until the EE
is commanded to move also in $x$-direction. Note that the pseudoinverse (\ref%
{J3R}) is the same when computed with an SVD or with the DLS method $\mathbf{%
J}_{\lambda }^{+}\left( \mathbf{q}_{0}\right) $ and setting $\lambda =0$.
The described phenomenon is not only problematic when the EE is commanded to
move along instantaneously infeasible directions, but generally when solving
the IK starting from a singularity.

A straightforward approach to tackle this issue is to perturb a singular
configuration $\mathbf{q}_{0}$ by a vector $\boldsymbol{\varepsilon }\in {%
\mathbb{R}}^{n}$ of small joint increments, expecting $\mathbf{J}\left( 
\mathbf{q}_{0}+\boldsymbol{\varepsilon }\right) $ to be non-singular (but
possibly ill-conditioned). Starting from $\mathbf{q}_{0}+\boldsymbol{%
\varepsilon }$, the IK solution is computed. This approach relies on a
proper choice of $\boldsymbol{\varepsilon }$. As the only purpose of $%
\boldsymbol{\varepsilon }$ is to get off the singularity, and since
singularities form a set of measure zero, random values could be used for $%
\boldsymbol{\varepsilon }$. Random $\boldsymbol{\varepsilon }$ will
unnecessarily deviate from the current $\mathbf{q}_{0}$, however. Perhaps
more critical is the fact that random values do not lead to a predetermined
behavior of the solution. Instead, an informed choice is needed.

In this paper an IK solution method is proposed where $\boldsymbol{%
\varepsilon }$ delivers only the motions necessary to reach a nearby regular
point. It employs results from the local analysis of singular motions, and
is therefore called \emph{Analytically Informed Inverse Kinematics Solution
(AI-IK)} method.

\section{Damped Least Squares (DLS) Solution\label{secDLS}%
\vspace{-2ex}%
}

To avoid large joint motions near singularities, in \cite%
{Wampler_DPI_1986,NakamuraHanafusa1986} IK solutions are sought such that $%
\left\Vert \mathbf{V}-\mathbf{J}\left( \mathbf{q}\right) \dot{\mathbf{q}}%
\right\Vert ^{2}+\lambda ^{2}\left\Vert \dot{\mathbf{q}}\right\Vert ^{2}$.
This leads to the damped least squares (DLS) solution $\dot{\mathbf{q}}=%
\mathbf{J}_{\lambda }^{+}\mathbf{V}$, with%
\begin{equation}
\mathbf{J}_{\lambda }^{+}=\mathbf{J}^{T}\left( \mathbf{JJ}^{T}+\lambda ^{2}%
\mathbf{I}\right) ^{-1}  \label{InvDamped}
\end{equation}%
where $\lambda $ is a damping factor. The form (\ref{InvDamped}) is
equivalent to the original expression $\mathbf{J}_{\lambda }^{+}=\left( 
\mathbf{J}^{T}\mathbf{J}+\lambda ^{2}\mathbf{I}\right) ^{-1}\mathbf{J}^{T}$.
Written in this way, it is obvious that $\mathbf{J}_{\lambda }^{+}$ becomes
the (non-weighted) right pseudoinverse when $\lambda =0$. It will therefore
be called the \emph{damped pseudoinverse (DPI)}. This form is also
numerically better conditioned.

The Jacobian is a $m\times n$ matrix of rank $r\leq \min \left( m,n\right) $%
. Using the singular value decomposition (SVD), the Jacobian is written as $%
\mathbf{J}=\mathbf{U}\boldsymbol{\Sigma }\mathbf{V}^{T}$, with $\boldsymbol{%
\Sigma }=\mathrm{diag~}\left( \sigma _{1},\ldots ,\sigma _{r},0,\ldots
,0\right) $, where $\sigma _{i}$ are the non-zero singular values of $%
\mathbf{J}$. The left pseudoinverse can then be expressed as $\mathbf{J}^{+}=%
\mathbf{V}\boldsymbol{\Sigma }^{+}\mathbf{U}^{T}$ with $\boldsymbol{\Sigma }%
^{+}=\mathrm{diag~}\left( \tfrac{1}{\sigma _{1}},\ldots ,\tfrac{1}{\sigma
_{r}},0,\ldots ,0\right) $. The DPI possesses the similar expression $%
\mathbf{J}_{\lambda }^{+}=\mathbf{V}\boldsymbol{\Sigma }_{\lambda }^{+}%
\mathbf{U}^{T}$, with $\boldsymbol{\Sigma }_{\lambda }^{+}=\mathrm{diag~}%
\left( \tfrac{\sigma _{1}}{\sigma _{1}^{2}+\lambda ^{2}},\ldots ,\tfrac{%
\sigma _{r}}{\sigma _{r}^{2}+\lambda ^{2}},0,\ldots ,0\right) $. For $%
\lambda \rightarrow 0$, the DPI becomes the left pseudoinverse $\mathbf{J}%
_{\lambda =0}^{+}=\mathbf{J}^{+}$. From this form it is clear that the
damping parameter $\lambda $ regularizes the solution. Even if $\mathbf{J}$
is regular, i.e. $r=m$, near singularities, the smallest singular value $%
\sigma _{r}$ becomes small, and the matrix to be inverted ill-conditioned.
This is avoided with sufficiently large $\lambda $. On the other hand, large 
$\lambda $ slow down the convergence of the iterative solution. Several
methods for adaptation of the damping factor $\lambda $ were proposed \cite%
{ChanLawrence1988,NakamuraHanafusa1986,MayorgaWongMilano1992,Chiaverini1997}.

In the expression for $\mathbf{J}$, the first $r$ columns of $\mathbf{U}%
\boldsymbol{\Sigma }$ form a basis for the image space of $\mathbf{J}$, and
the remaining columns (those multiplied with zero $\sigma _{r+1},\ldots
,\sigma _{m}$) span the space of twists that cannot be generated. Thus,
velocities $\mathbf{V}\in \ker \boldsymbol{\Sigma }_{\lambda }^{+}\mathbf{U}%
^{T}$ always lead to $\dot{\mathbf{q}}=\mathbf{0}$. Any form of
pseudoinverse method cannot solve the instantaneous IK problem when the
commanded EE-velocity can instantaneously not be generated by the
manipulator.

\section{Iterative Solution of Inverse Kinematics\label{secIK}%
\vspace{-2ex}%
}

Throughout the rest of this paper, $W=SE\left( 3\right) $ is assumed, for
simplicity. Let $\mathbf{q}\in {\mathbb{V}}^{n}$ be the joint coordinate
vector in the current pose, and $\mathbf{C}_{\mathrm{d}}\in SE\left(
3\right) $ the desired EE-pose. The 'difference' of the desired and the
current EE pose (i.e. the transformation from desired to current
configuration) is%
\begin{equation}
\Delta \mathbf{C}\left( \mathbf{q}\right) =\mathbf{C}^{-1}\left( \mathbf{q}%
\right) \mathbf{C}_{\mathrm{d}}.  \label{DeltaC}
\end{equation}%
Let $\mathbf{q}_{\mathrm{d}}\in {\mathbb{V}}^{n}$ be a (unknown) joint
coordinate vector corresponding to the desired $\mathbf{C}_{\mathrm{d}}$.
The IK problem is then to compute $\Delta \mathbf{q}_{\mathrm{d}}$ such that 
$\mathbf{q}_{\mathrm{d}}=\mathbf{q}+\Delta \mathbf{q}_{\mathrm{d}}$. For
small deviations, it is $\Delta \mathbf{C}-\mathbf{I}\in se\left( 3\right) $%
, and the first-order approximation of (\ref{DeltaC}) yields%
\begin{equation}
\Delta \mathbf{q=J}^{+}\left( \mathbf{q}\right) \Delta \mathbf{C}\left( 
\mathbf{q}\right) ^{\vee }  \label{Deltaq}
\end{equation}%
where $\mathbf{A}^{\vee }\in {\mathbb{R}}^{6}$ denotes the vector
corresponding to $\mathbf{A}\in se\left( 3\right) $. Here $\mathbf{J}^{+}$
is an appropriately chosen inverse, e.g. the DPI. The so obtained $\Delta 
\mathbf{q}$ will usually not solve the IK problem. Therefore, step (\ref%
{Deltaq}) is repeated with $\mathbf{q}$ replaced by $\mathbf{q}+\Delta 
\mathbf{q}$ until the error tracking satisfies $\left\Vert \Delta \mathbf{C}-%
\mathbf{I}\right\Vert _{\mathbf{M}}\leq \varepsilon $, where $\left\Vert
{}\right\Vert _{\mathbf{M}}$ is a left-invariant metric on $SE\left(
3\right) $. Notice that $\Delta \mathbf{C}-\mathbf{I}\approx \log \Delta 
\mathbf{C}$ is a first-order approximation of the logarithm on $SE\left(
3\right) $, i.e. $\Delta \mathbf{C}\left( \mathbf{q}\right) =\exp \left(
\Delta \mathbf{C}-\mathbf{I}\right) $ holds true for small deviations. If
the DPI is used in (\ref{Deltaq}), this resembles the Levenberg-Marquardt
method.

The solution (\ref{Deltaq}) can be complemented by a null-space solution in
case of redundant robots ($m<n$). The velocity command sent to the robot
controller is $\Delta \mathbf{q}/T$, with $T$ being the sampling time.

\section{Analytically Informed Numerical IK Solution\label{secRobustIK}%
\vspace{-2ex}%
}

\subsection{The Analytical Part ---Identifying Singular Motions%
\vspace{-2ex}%
}

In the analytic step, directions of motion transversal to the singular
motions are determined. This is pursued independently of the trajectory
tracking, which implies that singularities are known before task execution.

\begin{assumption}
The singular locus is either locally a manifold or it is the intersection of
smooth manifolds (e.g. bifurcations), i.e. there are smooth singular motions.
\end{assumption}

This assumptions is by no means restrictive as it is very rare that the
singular locus is not the intersection of lower-dimensional manifolds.
Tangents to finite singular motions can be determined via higher-order local
analysis \cite{JMR2018,CISMMueller2019}. Denote with $L_{m}=\left\{ \mathbf{q%
}\in {\mathbb{V}}^{n}|\mathrm{rank}~\mathbf{J}\left( \mathbf{q}\right)
<m\right\} $ the variety of singular motions. At $\mathbf{q}\in L_{m}$ the
kinematic tangent cone $C_{\mathbf{q}}^{\mathrm{K}}L_{m}\subset {\mathbb{R}}%
^{n}$ is the set of tangents to smooth finite singular motions through $%
\mathbf{q}$. The kinematic tangent cone is easily computed with a
higher-order local analysis using the instantaneous joint screws (columns of 
$\mathbf{J}$). Closed form and recursive algorithms were reported \cite%
{Mueller-MMT2019}. With the above assumption, the tangent cone $C_{\mathbf{q}%
}^{\mathrm{K}}L_{m}=K_{\mathbf{q}}^{m\left( 1\right) }\cup \cdots \cup K_{%
\mathbf{q}}^{m\left( \bar{m}\right) }$ is the union of $\bar{m}$ vector
spaces $K_{\mathbf{q}}^{m\left( i\right) },i=1,\ldots \bar{m}$ of dimensions
less than $n$. Denote with $\mathbf{s}_{1},\ldots ,\mathbf{s}_{\bar{s}}\in {%
\mathbb{R}}^{n}$ a set of $\bar{s}$ vectors that form a basis on $K_{\mathbf{%
q}}^{m\left( 1\right) }\cap \ldots \cap K_{\mathbf{q}}^{m\left( \bar{m}%
\right) }$, and introduce the matrix $\mathbf{S}=\left( \mathbf{s}%
_{1},\ldots ,\mathbf{s}_{\bar{s}}\right) \in {\mathbb{R}}^{n}$.

\subsection{The Numerical Part ---Regularizing the Jacobian%
\vspace{-2ex}%
}

The above discussion provides a means to solve the inverse kinematics at
singularities. To this end, introduce the projector $\mathbf{N}=\mathbf{SS}%
^{T}-\mathbf{I}$ to the orthogonal complement of $\mathbf{S}$. Let $%
\boldsymbol{\varepsilon }\in {\mathbb{R}}^{n}$ be an arbitrary vector. Its
projection $\mathbf{x}=\mathbf{N}\left( \mathbf{q}\right) \boldsymbol{%
\varepsilon }$ is transversal to singular motions through $\mathbf{q}$, and $%
\mathbf{x}$ represents tangents to motions away from the singular locus. For
sufficiently large $\boldsymbol{\varepsilon }$, the prolonged Jacobian $%
\mathbf{J}\left( \mathbf{q}_{0}+\mathbf{x}\right) $ is regular (recall
singularities form a lower-dimensional set in ${\mathbb{V}}^{n}$). The IK is
solved with $\mathbf{J}\left( \mathbf{q}_{0}+\mathbf{x}\right) $ in (\ref%
{Deltaq}), where $\mathbf{x}$ is determined with a predefined vector of
joint perturbations $\boldsymbol{\varepsilon }$. This method yields a motion
(according to prescribed $\boldsymbol{\varepsilon }$) for leaving the
singularities. The conditioning of $\mathbf{J}\left( \mathbf{q}_{0}+\mathbf{x%
}\right) $ depends on the magnitude of $\boldsymbol{\varepsilon }$. If
sufficiently large (e.g. in the order of the damping $\lambda ^{2}$), $%
\mathbf{J}\left( \mathbf{q}_{0}+\mathbf{x}\right) $ is regular, and a
non-regularized inverse is applicable.

In summary, the analytically informed inverse kinematics (AI-IK) method
consists in iteratively solving the IK problem starting with the
analytically regularized Jacobian $\mathbf{J}\left( \mathbf{q}_{0}+\mathbf{x}%
\right) $. In other words, the AI-IK method is the standard IK method where
the first iteration step is replaced by the regularizing increment $\mathbf{x%
}$.

\subsection{Analytic Calculation of the Prolonged Jacobian%
\vspace{-2ex}%
}

The Jacobian at the perturbed configuration can be expressed as $\mathbf{J}%
\left( \mathbf{q}_{0}+\mathbf{x}\right) =\mathbf{J}\left( \mathbf{q}%
_{0}\right) +d\mathbf{J}\left( \mathbf{q}_{0},\mathbf{x}\right) +\frac{1}{2}%
d^{2}\mathbf{J}\left( \mathbf{q}_{0},\mathbf{x}\right) +\ldots $, where the $%
k$th-order differential $d^{k}\mathbf{J}\left( \mathbf{q}_{0},\mathbf{x}%
\right) $ is a $k$th-degree monomial in $x_{i}$. The order up to which this
series must be written in order to obtain a regular Jacobian can be
determined exactly for a given robot. Column $i$ of $\mathbf{J}=\left( 
\mathbf{J}_{1},\ldots ,\mathbf{J}_{n}\right) $ is the instantaneous screw
coordinate vector $\mathbf{J}_{i}$ of joint $i$ expressed in EE-frame. Its
first differential is given as $d\mathbf{J}_{i}\left( \mathbf{q},\mathbf{x}%
\right) =\sum_{i<j\leq n}x_{j}\left[ \mathbf{J}_{i},\mathbf{J}_{j}\right] $,
where $\left[ \mathbf{J}_{i},\mathbf{J}_{j}\right] $ is the Lie bracket of
the two screws \cite{Mueller-MMT2019}. Thus $d\mathbf{J}\left( \mathbf{q}%
_{0},\mathbf{x}\right) $ involves the Lie brackets of all joint screws
multiplied with $x_{i}$, and the image of $\mathbf{J}\left( \mathbf{q}%
_{0}\right) +d\mathbf{J}\left( \mathbf{q}_{0},\mathbf{x}\right) $ is
contained in $\mathrm{span}\,\left( \mathbf{J}_{i},\left[ \mathbf{J}_{i},%
\mathbf{J}_{j}\right] \right) $, and its dimension is bigger than that of $%
\mathbf{J}\left( \mathbf{q}_{0}\right) $ provided that the $\left[ \mathbf{J}%
_{i},\mathbf{J}_{j}\right] \notin \mathrm{im}\,\mathbf{J}$ are not filtered
out by multiplication with $x_{i}$. The latter is avoided by using $\mathbf{x%
}$ transversal to singular motions, as determined above. This argument is
repeated for the second differential, which involves nested Lie brackets,
and so on for higher-order. It was shown that this bracketing process
terminates at a certain order with the smallest Lie algebra containing $%
\mathrm{im}\,\mathbf{J}$ for all $\mathbf{q}\in {\mathbb{V}}^{n}$ \cite%
{Mueller-MMT2019}, and this order is sufficient to generate a regular
Jacobian. Thus, accordingly truncating the series expansion in terms of $%
\mathbf{x}$ transversal to singular motions through $\mathbf{q}_{0}$ yields
a regularizing prolongation.

\section{Example: Redundant 7R Kuka LBR iiwa 14 R820%
\vspace{-2ex}%
}

The redundant 7 DOF Kuka iiwa robot serves as a realistic example. In the
reference configuration $\mathbf{q}_{0}{\mathbb{V}}^{7}$, shown in Fig. \ref%
{figExamples}b), the joint screw coordinates $\,\mathbf{J}_{i}\left( \mathbf{%
q}_{0}\right) $, deduced from the shown EE-frame, give rise to the Jacobian%
\begin{equation}
\mathbf{J}\left( \mathbf{q}_{0}\right) =\left( \mathbf{J}_{1},\ldots ,%
\mathbf{J}_{7}\right) ={\small \left( 
\begin{array}{ccccccc}
0 & 1 & 0 & 1 & 0 & 1 & 0 \\ 
0 & 0 & 0 & 0 & 0 & 0 & 0 \\ 
1 & 0 & 1 & 0 & 1 & 0 & 1 \\ 
0 & 0 & 0 & 0 & 0 & 0 & 0 \\ 
0 & -a-b-c & 0 & -b-c & 0 & -c & 0 \\ 
0 & 0 & 0 & 0 & 0 & 0 & 0%
\end{array}%
\right) .}  \label{J0}
\end{equation}%
The configuration is obviously singular since $\mathrm{rank}\,\mathbf{J}%
\left( \mathbf{q}_{0}\right) =3$ while the maximal rank in regular is $m=6$.
It was shown in \cite{JMR2018} that the tangent cone to the singular motions
through $\mathbf{q}_{0}$ is $C_{\mathbf{q}_{0}}^{\mathrm{K}}L_{6}=K_{\mathbf{%
q}_{0}}^{6\left( 1\right) }\cup K_{\mathbf{q}_{0}}^{6\left( 2\right) }$,
with $K_{\mathbf{q}_{0}}^{6\left( 1\right) }=\{\mathbf{x}\in {\mathbb{R}}%
^{7}|x_{4}=0\}$ and $K_{\mathbf{q}_{0}}^{6\left( 2\right) }=\{\mathbf{x}\in {%
\mathbb{R}}^{7}|x_{2}=0,x_{6}=0\}$. That is, $\mathrm{rank}\,\mathbf{J}<6$
as long as either joint 4 remains fixed or joints 2 and 6 remain locked.
Thus at $\mathbf{q}_{0}$ a 6-dim and a 5-dim variety of singularities
intersect, with respective tangent spaces $K_{\mathbf{q}_{0}}^{6\left(
1\right) }$ and $K_{\mathbf{q}_{0}}^{6\left( 2\right) }$. The detailed
analysis \cite{JMR2018} shows that this variety of singularities stratifies
in to a 3-dim manifold of rank 4, a 2-dim manifold of rank 3 singularities.
This is not important for solving the inverse kinematics which only needs to
ensure transversality to the singular variety.

The $\bar{s}=4$ independent vectors in $K_{\mathbf{q}_{0}}^{6\left( 1\right)
}\cap K_{\mathbf{q}_{0}}^{6\left( 2\right) }$ are%
\begin{eqnarray}
\mathbf{s}_{1} &=&\left( 1,0,0,0,0,0,0\right) ^{T},\mathbf{s}_{2}=\left(
0,0,1,0,0,0,0\right) ^{T}  \notag \\
\mathbf{s}_{3} &=&\left( 0,0,0,0,1,0,0\right) ^{T},\mathbf{s}_{4}=\left(
0,0,0,0,0,0,1\right) ^{T}.
\end{eqnarray}%
The projector $\mathbf{N}$, and the vector $\boldsymbol{\varepsilon }=\left(
\varepsilon _{1},\ldots ,\varepsilon _{7}\right) $ yield the vector of
regularizing joint perturbations $\mathbf{x=N}\boldsymbol{\varepsilon }%
=\left( 0,\varepsilon _{2},0,\varepsilon _{4},0,\varepsilon _{6},0\right)
^{T}$. In this singular configuration, first-order Lie brackets are
sufficient to generate the closure algebra from the above joint screws: $%
se\left( 3\right) =\mathrm{span}~\left( \mathbf{J}_{i},\left[ \mathbf{J}_{i},%
\mathbf{J}_{j}\right] \right) $. The first-order prolongated Jacobian is
thus $\mathbf{J}\left( \mathbf{q}_{0}+\mathbf{x}\right) =\mathbf{J}\left( 
\mathbf{q}_{0}\right) +d\mathbf{J}\left( \mathbf{q}_{0},\mathbf{x}\right) $,
with%
\begin{equation}
d\mathbf{J}\left( \mathbf{q}_{0},\mathbf{x}\right) ={\small \left( 
\begin{array}{ccccccc}
0 & 0 & 0 & 0 & 0 & 0 & 0 \\ 
0 & 0 & -\varepsilon _{2} & 0 & -\varepsilon _{24} & 0 & -\varepsilon _{246}
\\ 
0 & 0 & 0 & 0 & 0 & 0 & 0 \\ 
0 & 0 & -\varepsilon _{2}\left( a+b+c\right) & 0 & -a\varepsilon _{2}-\left(
b+c\right) \varepsilon _{24} & 0 & -a\varepsilon _{2}-b\varepsilon
_{24}-c\varepsilon _{246} \\ 
0 & 0 & 0 & 0 & 0 & 0 & 0 \\ 
0 & 0 & 0 & a\varepsilon _{2} & 0 & a\varepsilon _{2}+b\varepsilon _{24} & 0%
\end{array}%
\right) }  \label{dJ0}
\end{equation}%
where ${\varepsilon _{24}:=\varepsilon _{2}}+{\varepsilon _{4}}$, ${%
\varepsilon _{246}=\varepsilon _{2}}+{\varepsilon _{4}}+{\varepsilon _{6}}$.
Numerical results are computed with the geometric data $a=0.42,b=0.4,c=0.126$
(all in m, units omitted) for the parameters indicated in Fig. \ref%
{figExamples}b), according to the manufacturer data sheet. The desired
EE-pose is expressed relative to the EE-frame in the reference
configuration. The EE-pose is denoted as $\mathbf{C}=\left( \mathbf{R},%
\mathbf{r}\right) $ with rotation matrix $\mathbf{R}$ and position vector $%
\mathbf{r}$.%
\vspace{1ex}
\begin{figure}[b]
\includegraphics[width=11cm]{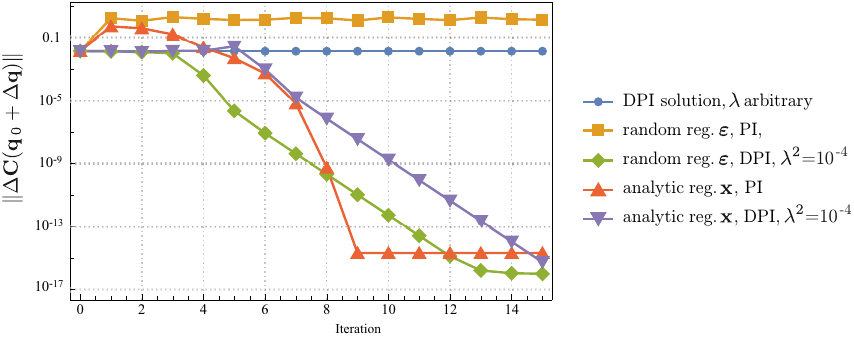} \hfill\ 
\caption{Convergence for 15 iterations. Notice that the error of the DPI
solution stays constant. The PI and DPI solution starting with the
analyticaly regularized configuration as well as the DPI starting with the
random pertrbation converge.}
\label{figXZCompare}
\end{figure}
\begin{figure}[bh]
a)\includegraphics[width=5.6cm]{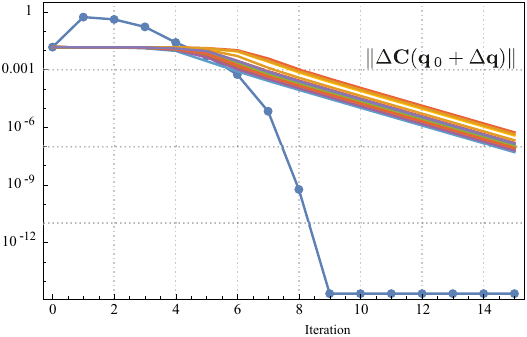}~~
b)\includegraphics[width=5.6cm]{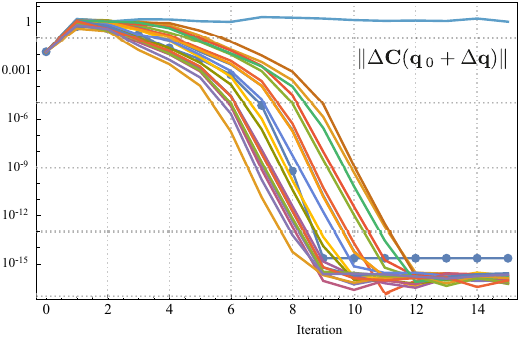}
\hfill\ 
\caption{Convergence of the DPI for 20 different random perurbations (curves
without markers), when a) $\protect\lambda ^{2}=10^{-4}$ and b) $\protect%
\lambda ^{2}=10^{-6}$. For comparison, the PI solution obtained with the
analyticaly regularized start configuration is included.}
\label{figXZRand}
\end{figure}

\textbf{1)} As first example, the extreme situation is considered when the
EE has to move in the $x$-$z$-plane, starting from the singular
configuration, according to $\mathbf{R}=\mathbf{I}$ and $\mathbf{r}=\left(
0.01,0,-0.01\right) ^{T}\,$m. As $\Delta \mathbf{C}^{\vee }=\left(
0,0,0,0.01,0,-0.01\right) ^{T}\in \ker \mathbf{J}^{T}\left( \mathbf{q}%
_{0}\right) =\ker \mathbf{J}_{\lambda }^{+}\left( \mathbf{q}_{0}\right) $,
the DPI solution (\ref{Deltaq}) yields $\Delta \mathbf{q=0}$, and the DLS
method cannot solve the IK problem. Moreover, since columns 2, 4, and 6 of $%
\mathbf{J}_{\lambda }^{+}\left( \mathbf{q}_{0}\right) $ are zero, any
instantaneous EE motion only involving translations along the $y$ and $z$
axis, or a rotation about the $x$-axis is completely annihilated.

The AI-IK method is applied with $\mathbf{x}=\left( 0,\varepsilon
_{2},0,\varepsilon _{4},0,\varepsilon _{6},0\right) ^{T}$ derived above
using $\varepsilon _{2}=\varepsilon _{4}=\varepsilon _{6}=10^{-3}$. This
perturbation leads to an initial error $\left\Vert \Delta \mathbf{C}\left( 
\mathbf{q}_{0}+\mathbf{x}\right) \right\Vert \dot{=}0.0145$. Since the robot
is redundant, the (undamped) right PI $\mathbf{J}_{\lambda =0}^{+}$ is used.
The convergence of the solution is shown in Fig. \ref{figXZCompare} for 15
iterations. Also shown is the result when using the DPI with $\lambda
^{2}=10^{-4}$.

For comparison, the IK is solved for 20 random perturbation vectors $%
\boldsymbol{\varepsilon }$, with $0\leq \varepsilon _{i}\leq 10^{-3}$, using
Mathematica (\texttt{RandomReal[0.001,7]}). The random numbers can be
reproduced by setting the seed value of the number generator to 12345 (%
\texttt{SeedRandom[12345]}). The convergence of the inverse kinematics
solution is shown in Fig. \ref{figXZCompare} when using the first seven
random numbers as elements of $\boldsymbol{\varepsilon }$. The DPI solution
converges while the undamped PI solution does not converge. To analyze the
performance of the IK solution with randomly perturbed initial
configuration, Fig. \ref{figXZRand} shows the DPI solutions for all 20
random vectors $\boldsymbol{\varepsilon }$. For damping $\lambda
^{2}=10^{-4} $ the solution converges for all 20 random vectors. A speed-up
of the slow convergence is achieved with a smaller damping. This is shown in %
\ref{figXZRand}b) for $\lambda ^{2}=10^{-6}$. However, the results show that
this may not be sufficient to cope with the ill-conditioned, and possibly
singular, Jacobian obtained with random perturbations.

\vspace{1ex}%

\begin{figure}[h]
\includegraphics[width=11cm]{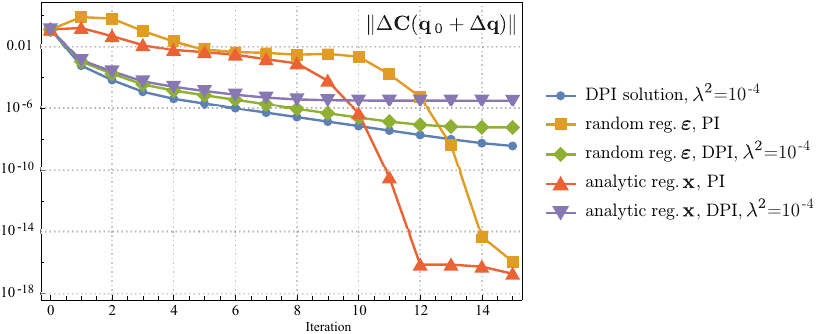}
\caption{Convergence during 15 iterations when perform a general motion
starting at the singularity.}
\label{figGeneral}
\end{figure}

\begin{figure}[h]
\includegraphics[width=10cm]{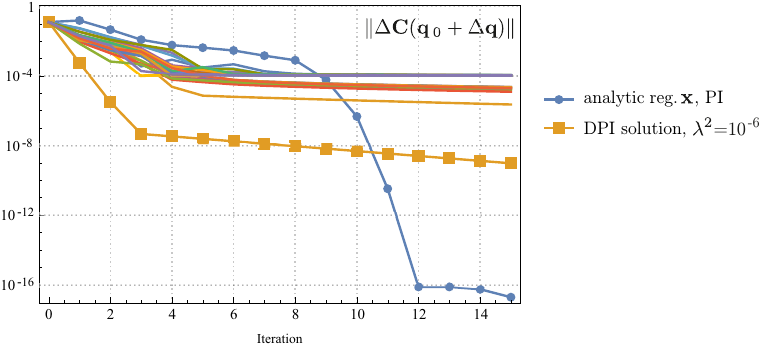}
\caption{Convergence of the DPI for 20 different random perurbations (curves
without markers), when $\protect\lambda ^{2}=10^{-6}$. For comparison, the
PI solution obtained with the analyticaly regularized start configuration,
and the DPI solution with $\protect\lambda ^{2}=10^{-6}$ are included.}
\label{figGeneralRandom}
\end{figure}

\textbf{2)} The EE is now following a general motion. The target EE pose at
the sampling step after the singularity is computed as $\mathbf{C}_{\mathrm{d%
}}=f\left( \mathbf{q}_{0}+\Delta \mathbf{q}_{\mathrm{d}}\right) $, with $%
\Delta \mathbf{q}_{\mathrm{d}}=\left(
0.01,0.01,0.05,0.01,0.01,0.01,0.05\right) ^{T}$. The convergence of the
iteration methods are shown in Fig. \ref{figGeneral}. The DPI method yields
a numerical solution, and exhibits the known convergence property for $%
\lambda ^{2}=10^{-4}$. The iteration starting with the analytically
regularized Jacobian and with the randomly perturbed Jacobian ($\mathbf{x}$
and $\boldsymbol{\varepsilon }$ as above) show similar convergence rates
when using the DPI. The undamped PI solutions achieve the most accurate
solutions. The convergence of the DPI solution, when starting from the
perturbed configurations according to the 20 different random vectors $%
\boldsymbol{\varepsilon }$, is shown in Fig. \ref{figGeneralRandom}.
Additionally, results of the DPI solution and the PI solution with
perturbation $\mathbf{x}$ (AI-IK solution) are shown again. 

From the examples it may be concluded that at singularities, the AI-IK
method should be used. This ensures that a solution of the IK problem will
always be found as this method is robust against the 'lock-up' phenomenon
observed in the first test situation. To further increase robustness, the
DPI of the regularized Jacobian may be used.

\section{Conclusion%
\vspace{-2ex}%
}

A lesser known problem of singular configurations of robotic manipulators is
that any PI-based IK solution method fails when instantaneous EE motions
necessary to reach a desired configuration is not feasible, i.e. the desired
EE-velocities are in the null-space of $\mathbf{J}^{T}$. To overcome this
problem, a solution method was proposed that is robust against such
situations. It relies on a perturbation away from the singularity, which is
obtained from a local analysis of the singular motions using the kinematic
tangent cone. The latter can be determined offline since the singularities
of the robot should be known. The method was shown for the classical
stretched arm singularities, but is applicable to general singularities.
Future work will address an informed choice of perturbation magnitude.
Another aspect to be addressed is how a preference for certain inverse
kinematics solution branches can be taken into account.

\begin{acknowledgement}
This work was supported by the LCM-K2 Center within the
framework of the Austrian COMET-K2 program, and by the
Austrian Science Fund (FWF) [I 4452-N]
\end{acknowledgement}

\bibliographystyle{spmpsci}
\bibliography{Fayet-Wolhart_ARK2020}

\end{document}